\documentclass[11pt]{article}

\usepackage[final]{acl}
\usepackage{multirow}
\usepackage{times}
\usepackage{latexsym}
\usepackage[T1]{fontenc}
\usepackage{amsmath}
\usepackage[utf8]{inputenc}
\usepackage{pifont}
\usepackage{microtype}
\usepackage{float}
\usepackage{inconsolata}
\usepackage{booktabs}
\usepackage{graphicx}
\usepackage{colortbl}
\newcommand{\pf}[1]{\cellcolor{red!30}#1}
\newcommand{\ps}[1]{\cellcolor{orange!30}#1}
\newcommand{\pt}[1]{\cellcolor{yellow!30}#1}

\DeclareRobustCommand\onedot{\futurelet\@let@token\@onedot}
\def\eg{\emph{e.g., }}  

\def\ie{\emph{i.e., }}

\title{AnimeAgent: Is the Multi-Agent via Image-to-Video models a Good Disney Storytelling Artist?}

\author{
 \textbf{Hailong Yan\textsuperscript{1,2}},
 \textbf{Shice Liu\textsuperscript{2}},
 \textbf{Tao Wang\textsuperscript{2}},
 \textbf{Xiangtao Zhang\textsuperscript{1}},
 \\
 \textbf{Yijie Zhong\textsuperscript{2}},
 \textbf{Jinwei Chen\textsuperscript{2}},
 \textbf{Le Zhang\textsuperscript{1,*}},
 \textbf{Bo Li\textsuperscript{2,*}}
\\
\\
 \textsuperscript{1}UESTC,
 \textsuperscript{2}vivo Mobile Communication Co. Ltd
\\
 \small{
   *~\textbf{Correspondence:} \href{lezhang@uestc.edu.cn}{lezhang@uestc.edu.cn}, \href{libra@vivo.com}{libra@vivo.com}
 }
}

\begin{document}

\maketitle

\begin{abstract}
Custom Storyboard Generation (CSG) aims to produce high-quality, multi-character consistent storytelling. Current approaches based on static diffusion models, whether used in a one-shot manner or within multi-agent frameworks, face three key limitations: (1) Static models lack dynamic expressiveness and often resort to ``copy-paste'' pattern. (2) One-shot inference cannot iteratively correct missing attributes or poor prompt adherence. (3) Multi-agents rely on non-robust evaluators, ill-suited for assessing stylized, non-realistic animation. To address these, we propose AnimeAgent, the first Image-to-Video (I2V)-based multi-agent framework for CSG. Inspired by Disney’s ``Combination of Straight Ahead and Pose to Pose'' workflow, AnimeAgent leverages I2V’s implicit motion prior to enhance consistency and expressiveness, while a mixed subjective-objective reviewer enables reliable iterative refinement. We also collect a human-annotated CSG benchmark with ground-truth. Experiments show AnimeAgent achieves SOTA performance in consistency, prompt fidelity, and stylization.
\end{abstract}

\section{Introduction}

Custom Storyboard Generation (CSG) aims to automatically produce storytelling images from character, scene, and script, reducing the high cost and repeated edits of manual drawing. Ideally, CSG should reflect animation’s distinctive dynamism and character interactions \cite{cartoonimator}.

\begin{figure}[h]
\centering
\includegraphics[width=\linewidth]{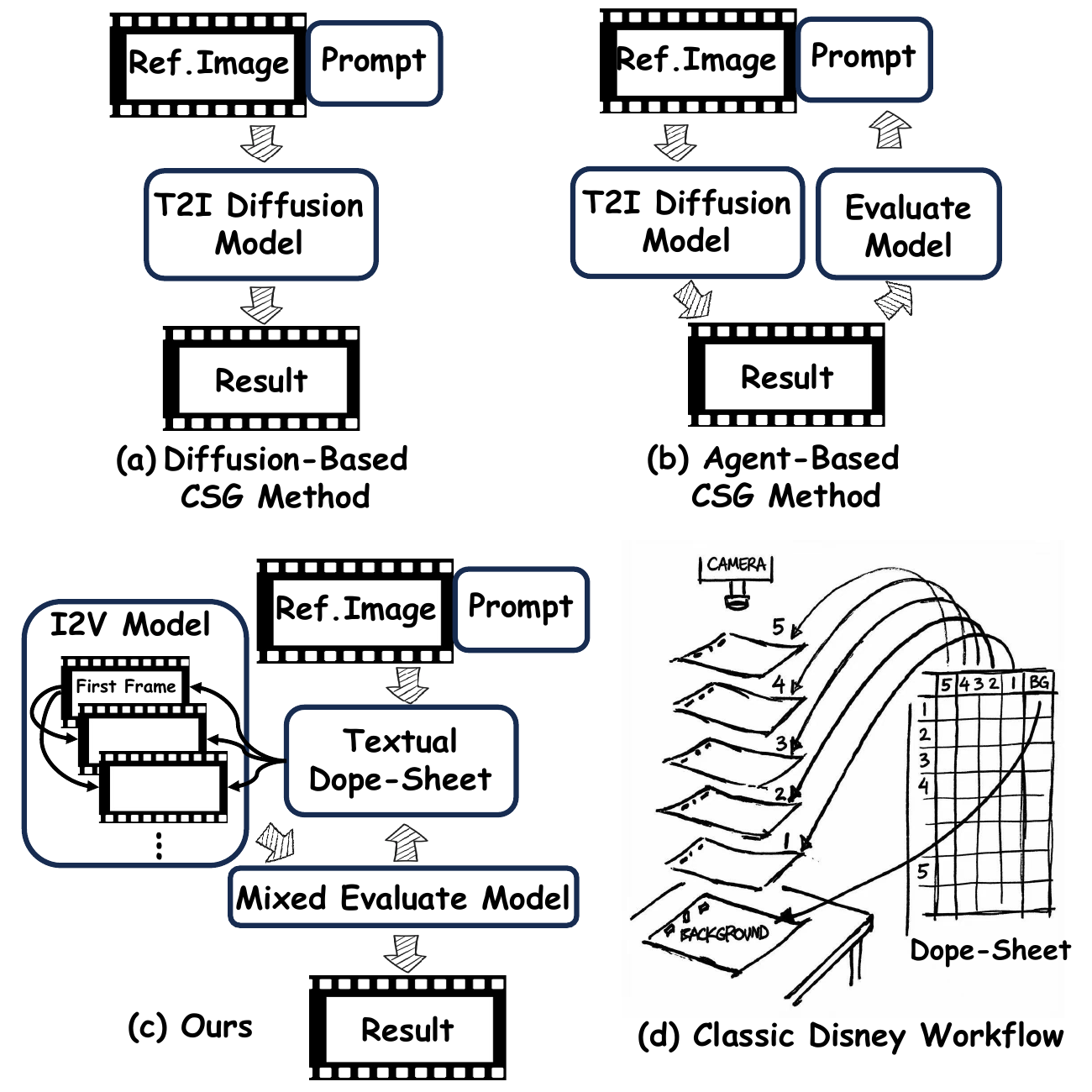}
\caption{\label{fig0}Comparison of paradigms for CSG. Our CSG paradigm is inspired by the ($d$).}
\vspace{-10pt}
\end{figure}

Early CSG methods train on fixed datasets to ensure consistency but generalize poorly to the new scene~\cite{seed}. To address this problem, some training-free approaches \cite{storydiffusion} improve generalization by sharing self-attention keys, values across frames without finetuning, yet often sacrifice prompt fidelity and struggle with multi-character interactions.

These methods all follow the one-shot static generation paradigm: ``one-shot'' collapses perception and synthesis into a single forward (Fig.~\ref{fig0}~(a)), lacking iterative refinement and often yielding missing attributes; ``static'' stems from the Text-to-Image (T2I) model’s lack of world knowledge, leading to repetitive ``copy-paste'' outputs that fail to render complex poses and multi-character interactions while preserving consistency \cite{cai2025diffusion}.

\begin{figure*}[h]
\centering
\includegraphics[width=\linewidth]{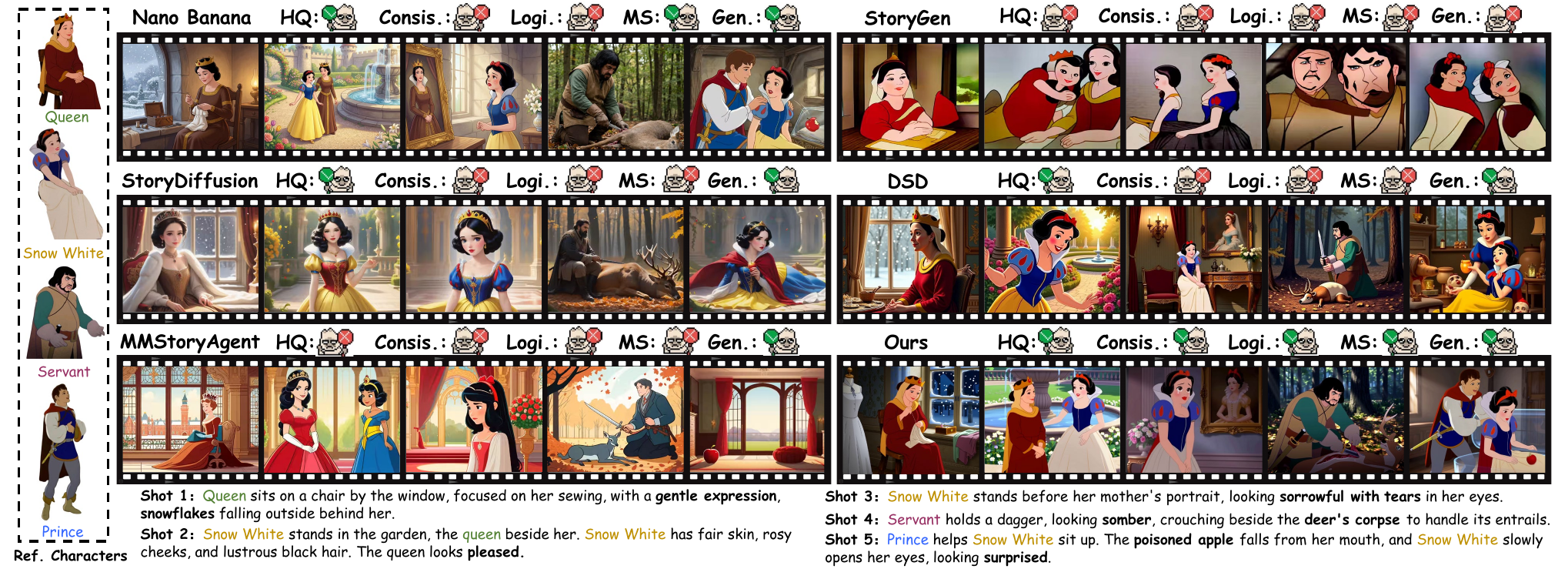}
\vspace{-10pt}
\caption{\label{fig1}Qualitative comparison of CSG methods across five criteria: High Quality, Consistency, Logical, Multi-Subjects, and Generalizability. Our method excels in visual fidelity, identity preservation, narrative coherence, and generalization — outperforming baselines that exhibit drift, incoherence, or stylistic inconsistency. }
\vspace{-10pt}
\end{figure*}

To incorporate feedback, recent work employs multi-agent frameworks built on T2I models to iteratively refine outputs \cite{xu2025mm} (Fig.~\ref{fig0}~(b)).
However, these approaches remain limited by two key issues: \ding{182} The generator is still a static T2I model; \ding{183} The evaluation relies on CLIP score or MLLM rewards \cite{xiang2025promptsculptor}. CLIP \cite{radford2021learning} is insensitive to artistic distortions and easily misled by visual anomalies, while MLLM scores are highly subjective and lack a stable reference \cite{li2024survey}.

To address these challenges, we propose AnimeAgent, a novel multi-agent framework, as shown in Fig.~\ref{fig0}~(c). Unlike prior T2I-based agent systems, AnimeAgent adopts dynamic modeling. Specifically, inspired by Disney’s \textit{``Combination of Straight Ahead and Pose to Pose''} workflow \cite{williams2012animator}, we reformulate CSG as a two-stage process: first \textbf{generating a continuous motion trajectory} and then \textbf{selecting the most expressive EXTREMES}. This paradigm naturally supports iterative refinement and mitigates limited world knowledge by capturing implicit motion dynamics.

Realizing this paradigm requires an engine that produces coherent motion trajectories. We find Image-to-Video (I2V) models provide an ideal foundation: their chain-of-frame inference not only overcomes the limitations of ``static'' but also offers two advantages in CSG:
\ding{182} First frame serves as a visual dope-sheet, initialized from a reference to stably anchor identity and style;
\ding{183} The temporal generation process acts as an implicit dynamic consistency constraint, preserving character identity and ensuring a plausible layout in complex scenes.

Due to a mismatch between user and model-preferred prompts, especially with ambiguous or simple prompt \cite{hei2024user}, which often directs I2V generation yields trajectories that deviate from narrative logic. To release I2V’s strengths, AnimeAgent employs a dual dope-sheet:the  Director parses the script and reference to produce a structured textual dope-sheet, while the I2V’s first frame acts as a visual dope-sheet, anchoring identity and style. Together, they enable semantics visual alignment, guiding I2V to generate trajectories that are both prompt faithful and animation expressive.

The trajectory is then evaluated by the Reviewer agent in two stages:
\ding{182} a consistency Reviewer detects contradictions in character identity, pose, or layout from fixed keyframes and feeds back to the Director to iteratively refine the textual dope sheet, addressing the lack of correction in one-shot generation;
\ding{183} a mixed Reviewer robustly assesses visual expressiveness by fusing automated metrics with human preferences, and selects the most dynamic Extremes as the final storytelling. Moreover, existing CSG benchmarks lack ground-truth annotations \cite{zhuang2025vistorybench}. To enable faithful and reliable evaluation, we construct a CSG reasoning dataset with human-annotated ground truth.

In summary, our contributions are: \ding{182} We propose AnimeAgent, the first I2V-based multi-agent framework for high-quality CSG. \ding{183} We introduce a hierarchical textual dope-sheet to unleash I2V’s potential and mixed reviewers for refinement and Extremes selection. \ding{184} We collect a CSG test set with ground-truth storyboards to support more realistic evaluation. \ding{185} AnimeAgent achieves SOTA performance, especially in consistency and prompt fidelity.

\begin{figure*}[htbp]
\centering
\vspace{-10pt}
\includegraphics[width=\linewidth]{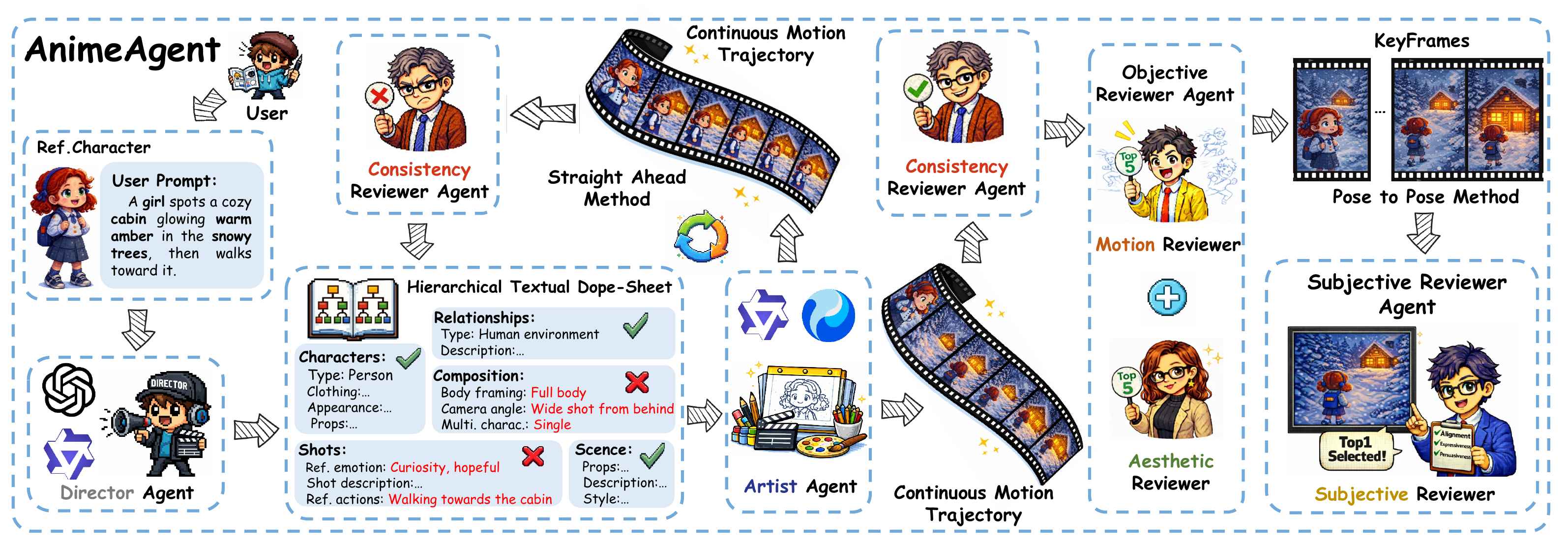}
\caption{\label{fig2}Overview of AnimeAgent. The Director Agent builds a hierarchical textual dope sheet from user inputs. The Artist Agent generates a continuous motion trajectory using an I2V model (“Straight Ahead”). A Consistency Reviewer validates keyframes for identity and layout, enabling refinement, while a Subjective Critic selects the most expressive “Extremes” via “Pose-to-Pose” evaluation, producing a coherent and visually compelling storytelling.}
\vspace{-8pt}
\end{figure*}

\section{Related Works}


\textbf{Train-based CSG methods}, initially built on GANs \cite{li2019storygan, li2022word}, used context encoders for inter-frame consistency but suffered from poor quality. Diffusion-based approaches \cite{maharana2022storydall, pan2024synthesizing} combine autoregressive modeling with historical image-text to enhance coherence. Recent works \cite{tao2024storyimager, zheng2025contextualstory} improve consistency further by replacing strict frame-by-frame generation with target-frame masks. However, these methods require large domain-specific datasets and generalize poorly to new characters or scenes.


\textbf{Train-free CSG methods} improve generalization by enhancing subject consistency through techniques such as self-attention \cite{storydiffusion, tewel2024training}, cross-attention \cite{cao2023masactrl}, feature injection \cite{tang2023emergent}, enhanced conditional prompting \cite{ye2023ip}, LLM-based prompt decomposition \cite{he2025dreamstory}, or unified prompt design \cite{liu2025one}, which applies a single prompt with dynamic sub-prompt reweighting. However, they often achieve only coarse identity alignment and can cause identity confusion in multi-character scenes.


\textbf{Agent-based CSG methods} move beyond one-shot T2I models to iterative, multi-agent frameworks that jointly model and character consistency. Recent approaches decouple narrative and generation \cite{zhou2025agentstory}, treat inconsistencies as repairable errors \cite{akdemir2025audit}, assign consistency as a dedicated agent role \cite{hu2024storyagent}, or employ a unified director agent \cite{li2024anim}. However, existing methods lack temporal motion modeling, resulting in limited dynamic coherence.

\section{AnimeAgent}
\subsection{Motivation and Overall Pipeline}


\begin{figure}[htbp]
\vspace{-10pt}
\centering
\includegraphics[width=\linewidth]{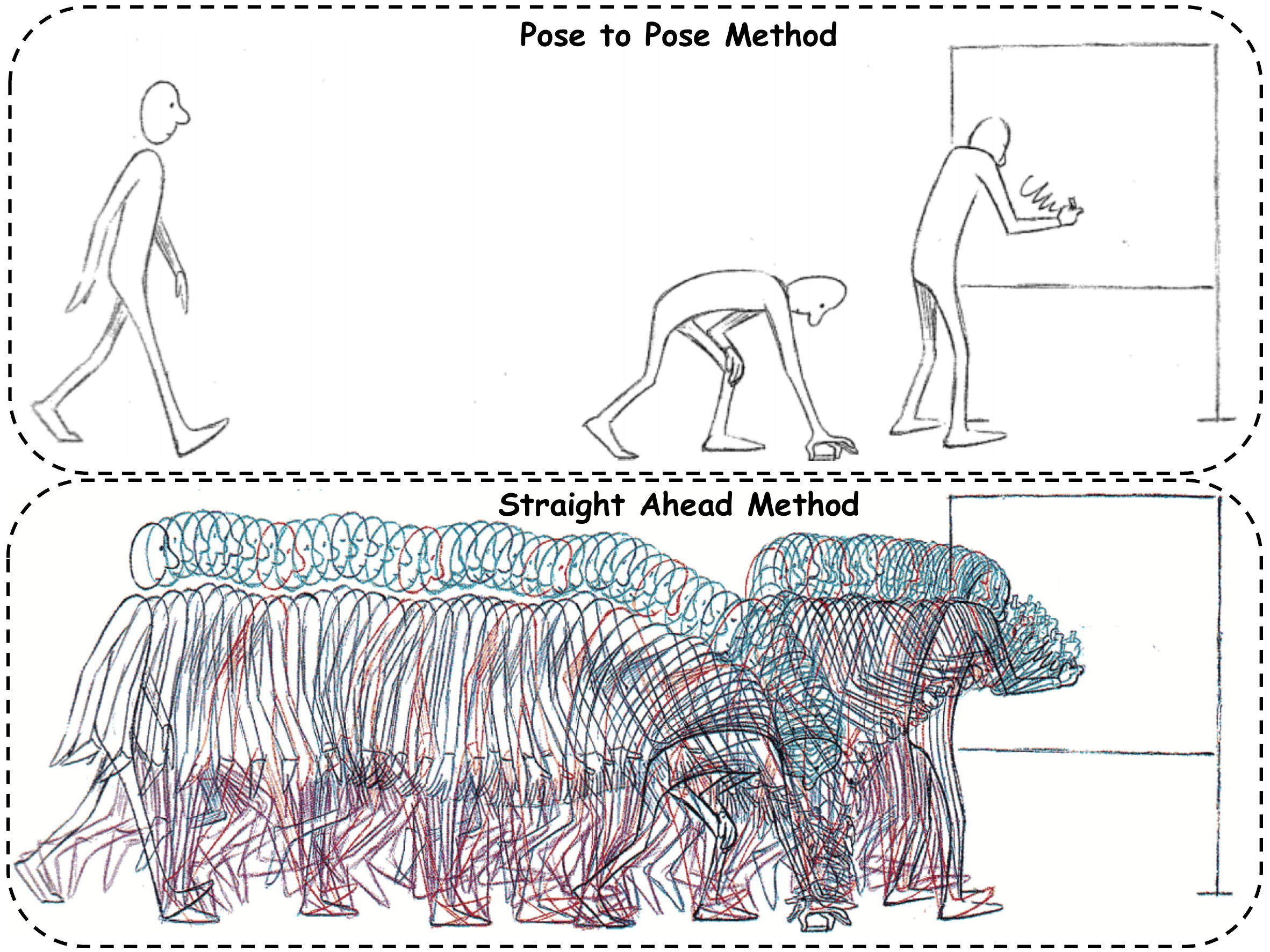}
\caption{\label{fig3}``Pose-to-Pose vs. Straight Ahead in Disney''}
\vspace{-12pt}
\end{figure}

For a story such as \textit{``a person walks to a blackboard, picks up chalk, and writes''} (Fig.~\ref{fig3}), animators typically combine two complementary workflows. \textbf{Pose-to-Pose} interpolates between key poses to ensure clear storytelling and logical composition, but it can feel rigid and less expressive. In contrast, \textbf{Straight Ahead} draws frames sequentially, producing fluid motion and spontaneous details, but it often sacrifices timing control. In practice, animators blend both: Pose-to-Pose for structure and Straight Ahead for vitality \cite{williams2012animator}. This blend provides a practical trade-off between narrative structure and expressive motion \cite{williams2012animator}.

To faithfully realize such stories while maintaining both timing control and expressive motion, we incorporate this mixed strategy into CSG, using the \textbf{Dope Sheet} (a structured script that encodes actions and shot content in traditional CSG) to specify actions and shot content. However, real-world CSG is not merely generative; it is inherently iterative. Animators repeatedly review, evaluate, and revise their work to refine motion and composition. This self-correcting process is crucial for complex actions, multi-character scenes, and stylized expression. To emulate this workflow, we introduce a multi-agent system called AnimeAgent that actively assesses and refines outputs, transforming CSG from one-pass generation into a reflective, iterative pipeline (Fig.~\ref{fig2}). The core components of our AnimeAgent are the Director Agent, the Artist Agent, and the Reviewer Agent. Thus, we introduce them as follows.



\subsection{Director Agent}


As shown in Fig.~\ref{fig2}, the Director Agent (D-A) translates non-expert inputs into a structured \textbf{textual dope sheet} (TS) that can be directly consumed by downstream I2V generation. Specifically, D-A takes a high-level story description \(T_{\text{ext}}\) as the primary input, and optionally leverages reference images for characters \(I_{\text{ref}}\) or scenes \(I_{\text{sce}}\). The resulting TS provides a precise and stable prompt interface, reducing ambiguity in free-form text and enabling controllable shot-wise synthesis.

\begin{figure}[t]
\vspace{-10pt}
\centering
\includegraphics[width=\linewidth]{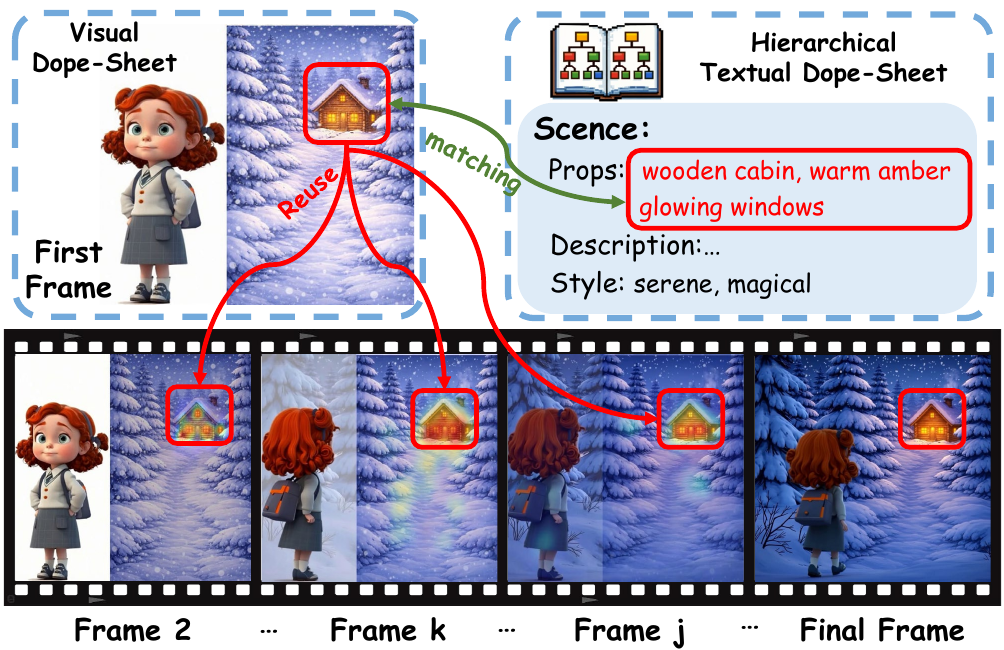}
\caption{\label{fig4} Text-Guided Semantic Anchor First Frame.}
\vspace{-10pt}
\end{figure}

To support controllable generation across multiple shots, TS is organized into five dimensions: \textbf{Characters}, \textbf{Shots}, \textbf{Scene}, \textbf{Composition}, and \textbf{Relationships}. Beyond per-shot specifications, D-A also introduces a binary linkage flag \(L \in \{T, F\}\) to enforce inter-shot consistency. When \(L = T\), the previous shot output \(\text{Res}_{j-1}\) is reused as the reference for the current shot, which helps preserve identity and appearance continuity during sequential generation.

D-A populates TS using multimodal large language models (MLLMs)~\cite{yang2025qwen3} by grounding each dimension in the available inputs. \textbf{Characters} are extracted from \(I_{\text{ref}}\) when provided, including entity type (\eg \textit{person}, \textit{animal}), clothing, appearance (\eg hair color, facial features), and props (\eg backpack). \textbf{Shots} are derived from \(T_{\text{ext}}\), specifying emotion (\textit{Ref.emo.}), narrative description (\textit{Desc.}), and action semantics (\textit{Ref.action}). The \textbf{Scene} dimension describes environment, key props, and artistic style; it fuses cues from \(I_{\text{sce}}\) with \(T_{\text{ext}}\) when \(I_{\text{sce}}\) is available, otherwise it relies on text only. Finally, \textbf{Composition} and \textbf{Relationships} are inferred jointly from \(T_{\text{ext}}\) and visual references: composition defines framing, camera angle, and multi-character layout, while relationships describe interaction types (\eg confrontation, collaboration) and their semantic details between characters or between characters and the environment.




\begin{figure}[t]
\vspace{-10pt}
\centering
\includegraphics[width=\linewidth]{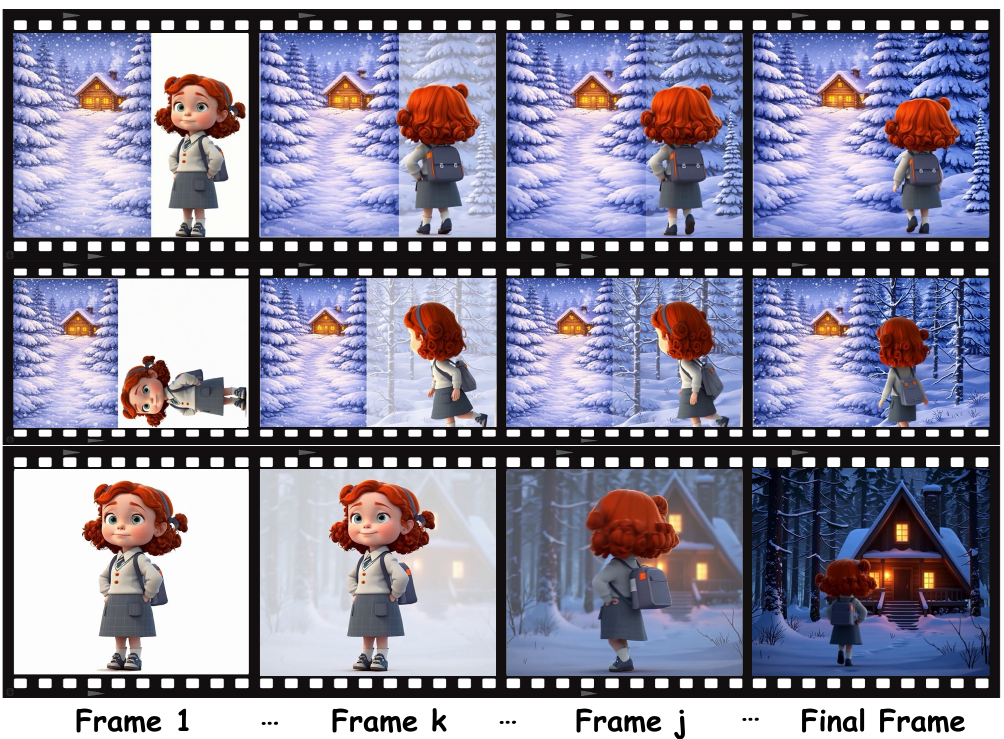}
\caption{\label{fig5}Layout-Robust Multi-Character I2V.}
\vspace{-10pt}
\end{figure}

\subsection{Artist Agent}

As shown in Fig.~\ref{fig2}, after the Director Agent produces the textual dope sheet (TS), we employ an Artist Agent to perform I2V-based generation and produce a continuous motion trajectory. Concretely, the Artist Agent uses an I2V model as its core engine, motivated by two key mechanisms: \ding{182} \textbf{The first frame as a visual dope sheet.} The first frame can anchor identity and key semantics, and these anchors can be propagated to subsequent frames. \ding{183} \textbf{Chain-of-frame generation for spatio-temporal reasoning.} Sequential generation implicitly supports spatio-temporal reasoning, enabling coherent layout and interaction modeling in complex multi-character scenes without explicit supervision.

\ding{182} Visual dope sheet effect. Although I2V models are trained on paired image-text data, we observe that under short or ambiguous prompts, cross-frame consistency becomes reliable only when the prompt explicitly specifies salient visual semantics that should be grounded in the first frame (\eg ``a wooden cabin with warm amber-lit windows''). As shown in Fig.~\ref{fig4}, the model sustains attention on the text-referenced regions of the first frame, which forms a closed-loop pathway from semantic specification to attention alignment and then to feature reuse. This indicates that the first frame is not merely an initialization, but a text-guided semantic anchor that enables stable feature propagation across the sequence.

\ding{183} Intrinsic spatio-temporal reasoning.
Building on \ding{182}, we further ask: \textit{Does the spatial positioning in the first frame affect the final sequence?} As shown in Fig.~\ref{fig5}, when we reposition characters and scene elements (\eg placing the girl left/right/above/below the cabin), the I2V model still generates coherent motion trajectories with correct identity, action progression, and spatial logic. This reveals a deeper capability beyond feature anchoring: the model can infer spatial and motion relationships from context and dynamically construct plausible interactions to support multi-character scenes. Compared with static T2I models that tend to rigidly copy composition, I2V treats the first frame as a semantic seed and maintains consistent multi-entity reasoning through temporal generation.

\begin{table*}[!ht]
    \centering
    \setlength{\arrayrulewidth}{1.2pt} 
    \resizebox{\textwidth}{!}{%
    \begin{tabular}{c|cc|cc|ccccc|c|c}
    \hline
        \multirow{2}{*}{Methods} &
        \multicolumn{2}{c|}{CSD$\uparrow$} &
        \multicolumn{2}{c|}{CIDS$\uparrow$} &
        \multicolumn{5}{c|}{PA$\uparrow$} &
        \multirow{2}{*}{CM$\uparrow$} &
        \multirow{2}{*}{Aes$\uparrow$} \\ \cline{2-10}
        & Cross & Self & Cross & Self & Scene & Shot & CI & IA & Avg. & & \\ \hline
        StoryGen \cite{intelligent} & 0.371 & 0.531 & 0.417 & 0.568 & 1.26 & 2.29 & 1.42 & 1.71 & 1.67 & 50.8 & 4.02 \\
        TheaterGen \cite{cheng2024theatergen}& 0.184 & 0.392 & 0.348 & 0.578 & 2.76 & 1.81 & 0.93 & 1.06 & 1.64 & 55.4 & 4.90 \\
        StoryDiffusion \cite{storydiffusion}& 0.340 & 0.547 & 0.436 & 0.565 & 2.00 & \ps{3.45} & 2.64 & 2.68 & 2.69 & 57.4 & 5.13 \\
        Anim-Director \cite{li2024anim}& 0.288 & 0.510 & 0.401 & 0.578 & \pt{3.64} & 3.07 & \pt{3.32} & \pt{2.69} & \pt{3.18} & 67.4 & \pt{5.59} \\
        Story-Adapter \cite{mao2024story} & 0.325 & \pt{0.737} & 0.401 & 0.626 & 1.93 & 3.28 & 2.55 & 2.46 & 2.56 & 59.3 & 4.89 \\
        SEED-Story \cite{seed}& 0.227 & \ps{0.748} & 0.287 & 0.587 & 1.96 & 1.83 & 0.46 & 0.67 & 1.23 & 44.4 & 3.82 \\
        UNO \cite{wu2025less}& 0.391 & 0.602 & 0.485 & 0.620 & 3.51 & 3.25 & 2.79 & 2.48 & 3.01 & \ps{74.2} & 5.23 \\
        OmniGen2 \cite{wu2025omnigen2}& \ps{0.454} & 0.600 & \pt{0.548} & \pt{0.647} & 3.48 & \pt{3.43} & 2.87 & 2.61 & 3.10 & \pt{70.2} & 5.25 \\
        CharaConsist \cite{wang2025characonsist}& 0.282 & 0.553 & 0.315 & 0.519 & 3.58 & 3.39 & 2.76 & 2.23 & 2.99 & 57.9 & \pf{5.88} \\
        Qwen-ImageEdit \cite{wu2025qwen} & 0.381 & 0.593 & 0.475 & 0.574 & \pf{3.91} & 3.23 & \ps{3.44} & 2.54 & \ps{3.28} & 59.8 & 5.50 \\
        MM-StoryAgent \cite{xu2025mm}& 0.238 & 0.669 & 0.388 & 0.596 & 2.92 & 2.61 & 1.85 & 1.63 & 2.25 & 61.5 & \pf{5.88} \\
        FreeCus \cite{zhang2025freecus} & 0.391 & 0.510 & 0.489 & 0.542 & 3.11 & 3.05 & 2.59 & 2.64 & 2.85 & 58.2 & 5.13 \\
        DSD \cite{cai2025diffusion}& \pt{0.417} & 0.544 & \ps{0.607} & \ps{0.688} & 3.04 & 3.15 & 2.67 & \ps{2.91} & 2.94 & 64.9 & 5.57 \\
        \hline
        Ours & \pf{\textbf{0.616}} & \pf\textbf{{0.785}} & \pf{\textbf{0.727}} & \pf{\textbf{0.767}} & \ps{\textbf{3.73}} & \pf{\textbf{3.85}} & \pf{\textbf{3.81}} & \pf{\textbf{3.04}} & \pf{\textbf{3.61}} & \pf{\textbf{77.6}} & \ps{\textbf{5.74}} \\ \hline
    \end{tabular}%
    }
    \caption{Quantitative Results of Various CSG Methods on ViStoryBench \cite{zhuang2025vistorybench}. CI: Character Interaction, IA: Individual Action; The \colorbox{red!40}{\strut first}, \colorbox{orange!50}{\strut second}, and \colorbox{yellow!50}{\strut third} values are highlighted.}
    \label{tab1}
    \setlength{\arrayrulewidth}{0.4pt}
\end{table*}

\subsection{Reviewer Agent}
Following the Director Agent that produces the textual dope sheet (TS) and the Artist Agent that generates an initial continuous motion trajectory, we introduce the Reviewer Agent to close the loop via evaluation and refinement. The Reviewer Agent is motivated by two key insights: \ding{182} The final frame in video generation results often fails to capture the core narrative intent. \ding{183} Existing rewards fail to reliably assess narrative quality and dynamic expressiveness in CSG.



For \ding{182}, inspired by Disney’s ``Pose-to-Pose '', narrative tension lies in EXTREMES that expressive frames capturing peak motion, not static endpoints. As Fig.~\ref{fig3} shows, the final stage of ``writing at a blackboard'' misses the key action ``picking up chalk.'' We thus shift evaluation from completion states to moments that best convey the plot.

For \ding{183}, existing agent-based CSG methods rely on CLIPScore \cite{xiang2025promptsculptor}, which is insensitive to exaggerated poses and non-photorealistic styles, or on direct MLLM ratings \cite{liu2025uve}, which are subjective and unstable, resulting in ineffective refinement feedback.

Stage 1: Iterative Refinement. The Consistency Reviewer leverages the MLLM as a descriptor. It uniformly samples five key frames from the tail of the video result and prompts the MLLM to produce a Textual Dope-Sheet (TS$_{\text{caption}}$). Using an embedding model ($e$), it encodes the textual entry of each dimension ($d$) in TS$_{\text{cap.}}$, TS$_{\text{ori.}}$ and computes cosine similarity as the consistency score (CS$_{d}$).
\vspace{-10pt}
\begin{equation}
\text{CS}_d = \frac{
    \mathbf{e}\big( [TS_{\text{cap.}}]_d \big) \cdot 
    \mathbf{e}\big( [TS_{\text{ori.}}]_d \big)
}{
    \left\| \mathbf{e}\big( [TS_{\text{cap.}}]_d \big) \right\| \,
    \left\| \mathbf{e}\big( [TS_{\text{ori.}}]_d \big) \right\|
}.
\end{equation}

Dimensions scoring below 0.8 trigger localized updates to their respective T-S entries, with at most two rounds to balance efficiency and fidelity.

\begin{figure}[t]
\vspace{-10pt}
\centering
\includegraphics[width=\linewidth]{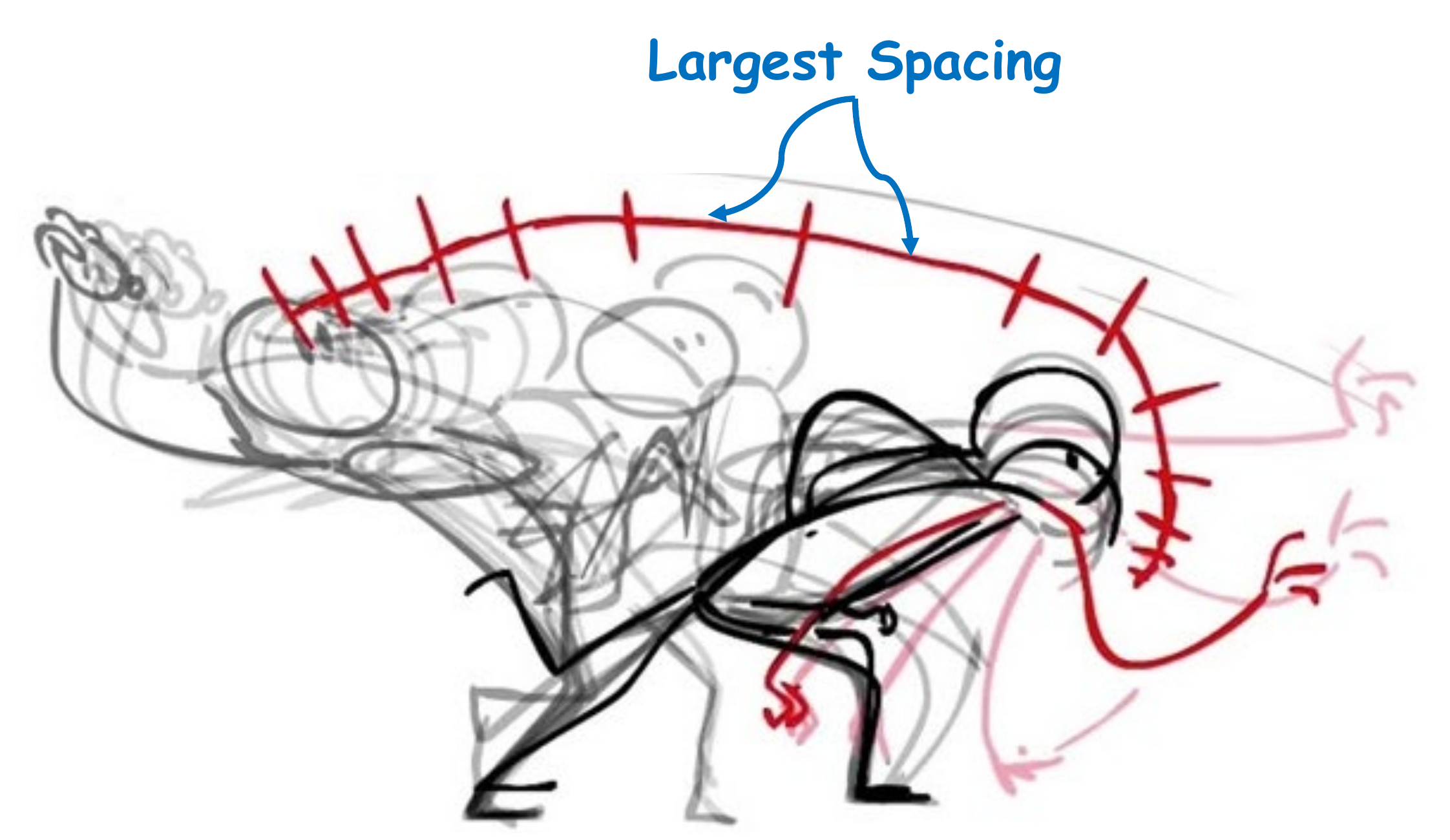}
\caption{\label{fig6}Identifying EXTREMENS via Motion Score.}
\vspace{-15pt}
\end{figure}

\begin{table*}[t]
    \centering
    \setlength{\arrayrulewidth}{1.2pt}
    \resizebox{\textwidth}{!}{%
    \begin{tabular}{c|cc|cc|ccccc|c|c}
    \hline
        \multirow{2}{*}{Methods} &
        \multicolumn{2}{c|}{CSD$\uparrow$} &
        \multicolumn{2}{c|}{CIDS$\uparrow$} &
        \multicolumn{5}{c|}{PA$\uparrow$} &
        \multirow{2}{*}{CM$\uparrow$} &
        \multirow{2}{*}{Aes$\uparrow$} \\ \cline{2-10}
        & Cross & Self & Cross & Self & Scene & Shot & CI & IA & Avg. & & \\ \hline
        MOKI~\cite{moki2024}& 0.214 & 0.694 & 0.372 & 0.621 & 1.88 & 0.94 & 2.88 & 1.10 & 1.70 & 46.0 & \pf{\textbf{5.79}} \\
        MorphicStudio~\cite{morphic2024studio}& \ps{0.577} & 0.628 & \ps{0.603} & \pt{0.677} & 3.07 & 2.66 & 3.36 & 2.11 & 2.80 & 60.8 & 4.96 \\
        Aibrm~\cite{brmgo2025} & 0.412 & \ps{0.730} & \pt{0.557} & \ps{0.740} & 3.05 & 2.53 & 3.56 & 2.05 & 2.80 & \ps{75.5} & \pt{5.72} \\
        ShenBi~\cite{shenbi2025} & 0.275 & 0.575 & 0.418 & 0.585 & 2.81 & 3.45 & 3.79 & 2.72 & 3.19 & 61.3 & 5.07 \\
        Typemovie~\cite{typemovie2024} & 0.325 & 0.646 & 0.464 & 0.621 & 3.00 & 2.61 & 2.86 & 2.30 & 2.69 & \pt{74.1} & 5.32 \\
        Doubao~\cite{doubao2024} & 0.367 & \pt{0.695} & 0.446 & 0.642 & 3.23 & 3.77 & \ps{3.98} & \pt{2.95} & 3.48 & 65.2 & 5.61 \\
        GPT4o~\cite{hurst2024gpt4o} & \pt{0.481} & 0.680 & 0.420 & 0.522 & \ps{3.65} & \pt{3.78} & \pt{3.91} & 2.68 & \pt{3.51} & 69.3 & 5.49 \\
        Gemini2.5~\cite{google2025nanobanana} & 0.447 & 0.657 & 0.553 & 0.642 & 3.32 & 3.71 & \pf{3.99} & \pf{3.25} & \ps{3.57} & 64.9 & 5.61 \\
        SeedDream-4.0~\cite{gong2025seedream} & 0.369 & 0.585 & 0.280 & 0.539 & \pt{3.43} & \ps{3.82} & \ps{3.98} & 0.53 & 2.94 & 49.5 & 5.21 \\
        \hline
        Ours & \pf{\textbf{0.616}} & \pf\textbf{{0.785}} & \pf{\textbf{0.727}} & \pf{\textbf{0.767}} & \pf{\textbf{3.73}} & \pf{\textbf{3.85}} & \textbf{3.81} & \ps{\textbf{3.04}} & \pf{\textbf{3.61}} & \pf{\textbf{77.6}} & \ps{\textbf{5.74}} \\ \hline
    \end{tabular}%
    }
        \caption{Quantitative Results of Various Commercial Platforms on ViStoryBench \cite{zhuang2025vistorybench}. }
    \label{tab2}
    \setlength{\arrayrulewidth}{0.4pt}
\end{table*}

\begin{table*}[!ht]
    \centering
    \setlength{\arrayrulewidth}{1.2pt}
    \resizebox{\textwidth}{!}{%
    \begin{tabular}{c|cc|cc|ccccc|c|c|c}
    \hline
        \multirow{2}{*}{Methods} &
        \multicolumn{2}{c|}{CSD$\uparrow$} &
        \multicolumn{2}{c|}{CIDS$\uparrow$} &
        \multicolumn{5}{c|}{PA$\uparrow$} &
        \multirow{2}{*}{CM$\uparrow$} &
        \multirow{2}{*}{Aes$\uparrow$} &
        \multirow{2}{*}{CLIP-I$\uparrow$} \\ 
        \cline{2-10}
        & Cross & Self & Cross & Self & Scene & Shot & CI & IA & Avg. & & &\\ \hline
        StoryDiffusion \cite{storydiffusion} & 0.422 & 0.485 & 0.443 & 0.461 & 2.26 & 2.77 & 2.59 & 2.45 & 2.52 & 52.6 & 5.14 & 0.615 \\
        Anim-Director \cite{li2024anim} & 0.396 & 0.431 & 0.487 & 0.443 & 3.16 & \pt{3.27} & 2.75 & 2.85 &  3.01 & 54.6 & \ps{5.48} & 0.585\\
        OmniGen2 \cite{wu2025omnigen2} & \pt{0.515} & 0.628 & \ps{0.627} & 0.591 & \ps{3.45} & 2.97 & 2.83 & 2.72 & 2.99 & 58.9 & 5.23 & 0.704\\
        FreeCus \cite{zhang2025freecus} & 0.474 & 0.538 & 0.531 & 0.576 & 2.95 & 2.86 & 2.44 & 2.52 & 2.69 & 55.3 & 5.08 & 0.662 \\
        Qwen-ImageEdit \cite{wu2025qwen} & \ps{0.532} & \ps{0.645} & \pt{0.611} & \ps{0.628} & \pt{3.36} & \ps{3.32} & \ps{3.45} & \pt{2.91} & \ps{3.26} & \ps{67.5} & \pt{5.41} & \ps{0.751}\\
        DSD \cite{cai2025diffusion} & 0.501 & \pt{0.637} & 0.584 & \pt{0.602} & 3.14 & 3.12 & \pt{3.34} & \ps{2.98} & \pt{3.15} & \pt{62.1} & 5.34 & \pt{0.712}\\
        \hline
        Ours & \pf{\textbf{0.583}} & \pf\textbf{{0.674}} & \pf{\textbf{0.655}} & \pf{\textbf{0.651}} & \pf{\textbf{3.55}} & \pf{\textbf{3.61}} & \pf{\textbf{3.53}} & \pf{\textbf{3.01}} & \pf{\textbf{3.43}} & \pf{\textbf{71.2}} & \pf{\textbf{5.53}} & \pf{\textbf{0.786}} \\ \hline
    \end{tabular}%
    }
        \caption{Quantitative Results of Various CSG Methods on AnimeBoard-GT.}
    \label{tab3}
    \setlength{\arrayrulewidth}{0.4pt}
\end{table*}

Stage 2: To identify the best extremes, we adopt a mixed reviewer framework, where each component addresses a distinct aspect of CSG quality.

The Objective Reviewer evaluates frame-level properties that are directly observable in the visual content:
(1) Aesthetic score (AES), evaluated by a pretrained aesthetic model \cite{zhang2024learning};
(2) Motion score (MOS). Optical flow–based method \cite{teed2020raft} often confuse camera motion with foreground dynamics. Unlike this, MOS first establishes keypoints between the reference and the I2V’s first frame using LightGlue \cite{lindenberger2023lightglue}, then tracks these points across frames with CoTracker \cite{karaev2025cotracker3}. The average displacement magnitude between each frame and its temporal neighbors is computed as a proxy for dynamic intensity, as shown in Fig.~\ref{fig6}. 
\vspace{-6pt}
\begin{equation}
    \text{MOS}_t = \frac{1}{N_t} \sum_{i=1}^{N_t} \| \mathbf{p}_i^t - \mathbf{p}_i^{t-1} \|,
    \label{eq2}
\end{equation}
where $\mathbf{p}_i^t$ denotes the 2D coordinates of the $i_{th}$ tracked keypoint in frame $t$, and $N_t$ is the number of successfully tracked keypoints in that frame. These two objective scores jointly select the top-5 candidate frames: $S_{\text{obj}} = \frac{\text{AES} + \text{MOS}}{2}.$

Complementing this, the Subjective Reviewer assesses narrative dimensions implicit in the script but not visible in pixels, such as intent, emotion, and storytelling efficacy. An MLLM conducts a structured triadic evaluation along three axes: \ding{182} \textbf{Alignment:} Does the frame faithfully convey the script’s implicit plot, motivation, or emotion? \ding{183} \textbf{Expressiveness:} Does it leverage CSG-specific techniques to amplify storytelling? \ding{184} \textbf{Persuasiveness:} Can it elicit viewer comprehension and emotional resonance?  And each axis is scored on a 0-5 integer scale based on the MLLM’s analysis, and the final subjective score is: $S_{\text{subj}} = \frac{\text{Alig.} + \text{Expr.}+ \text{Pers.}}{3}.$





\begin{figure*}[htbp]
\centering
\vspace{-2pt}
\includegraphics[width=\linewidth]{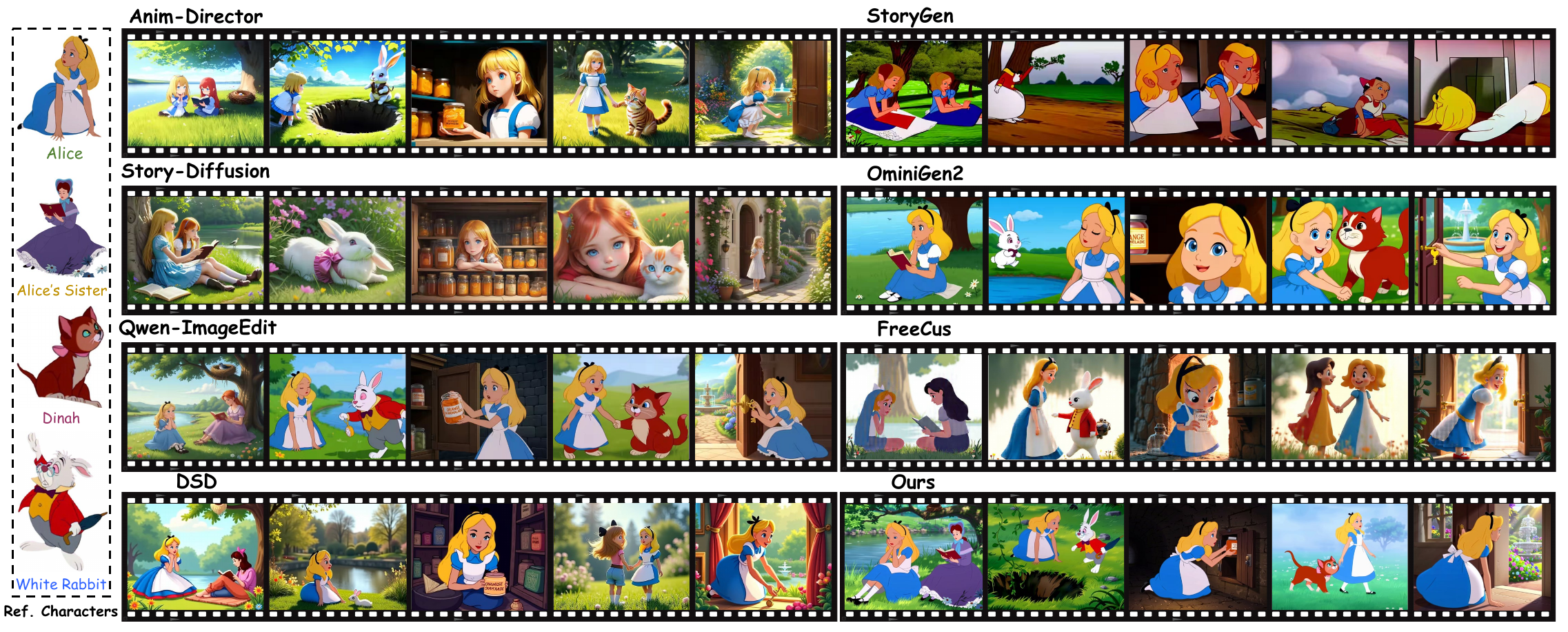}
\caption{\label{fig7}Qualitative Comparison Across Methods on Alice Story.}
\vspace{-10pt}
\end{figure*}

\begin{figure*}[htbp]
\centering
\vspace{-2pt}
\includegraphics[width=\linewidth]{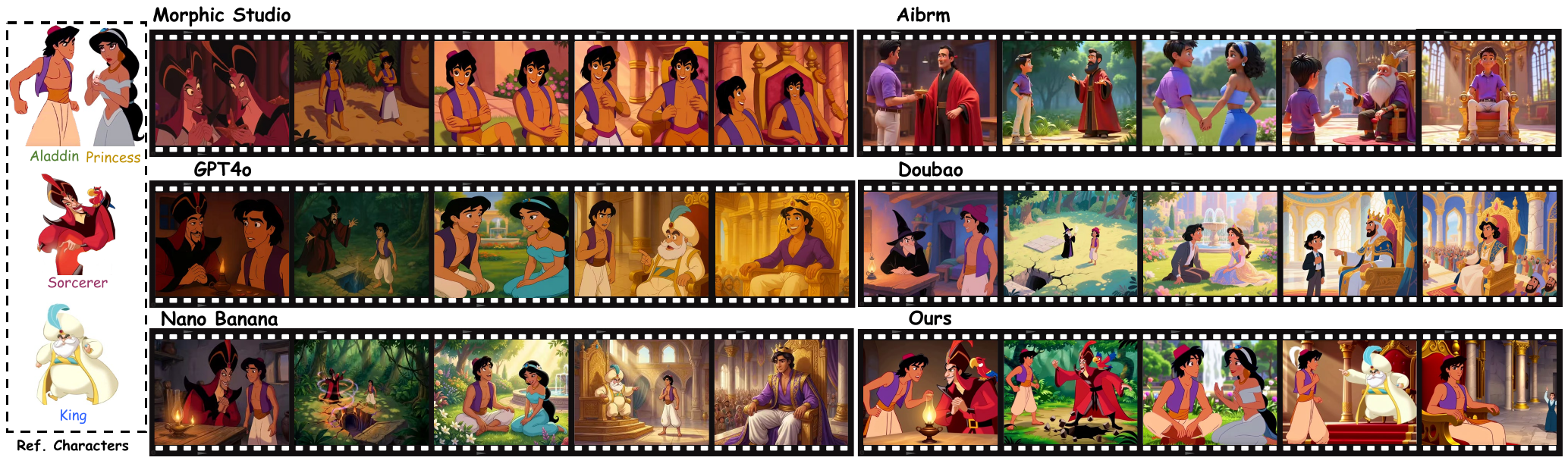}
\caption{\label{fig8}Qualitative Comparison Across Methods on Aladdin Story.}
\vspace{-10pt}
\end{figure*}

\section{Experiments}
\subsection{Implementation} 




We use Qwen3VL-8B \cite{yang2025qwen3} as the Director and Subjective Reviewer, Wan2.2-I2V-14B \cite{wan2025wan} as the Artist, and Qwen3-Embedding-0.6B \cite{zhang2025qwen3} as the embedding model for the Consistency Reviewer. Prompt examples are in the \textbf{Appendix}. We evaluate on the VistoryBench \cite{zhuang2025vistorybench} and our proposed AnimeBoard-GT, which includes human-annotated ground-truth storytellings for comprehensive evaluation. Following the VistoryBench, we compare AnimeAgent against 13 CSG methods and 9 commercial platforms. Following the VistoryBench metrics: Style Similarity (CSD), Character Identity Similarity (CIDS), Prompt Alignment (PA), Character Matching (CM), and Aesthetic Score (Aes). CLIP-I denotes similarity between the results and the GT.


\subsection{Comparison Results}





\textbf{Quantitative results.} We evaluate AnimeAgent under three complementary settings (Tables~\ref{tab1}--\ref{tab3}): (1) against CSG-based methods on ViStoryBench; (2) against commercial platforms on the same benchmark; and (3) on our new AnimeBoard-GT dataset with GT. Specifically, in Table~\ref{tab1}, AnimeAgent outperforms all baselines on every metric. Strong Style Similarity (CSD: 0.616/0.785) and Character Identity Similarity (CIDS: 0.727/0.767) indicate that I2V-based temporal modeling reduces identity drift and style fragmentation, while the best Prompt Alignment (Avg. 3.61) reflects accurate execution of complex narrative instructions. In Table~\ref{tab2}, commercial platforms (\eg Aibrm) achieve competitive aesthetics (Aes: 5.8) but fall behind in prompt alignment (2.80 vs. 3.61), often producing visually pleasing yet semantically off-target results. AnimeAgent preserves high aesthetics while improving semantic fidelity and character consistency. In Table~\ref{tab3}, AnimeAgent achieves the best CLIP-I (0.786) on AnimeBoard-GT, demonstrating strong generalization and alignment with GT content.

\begin{figure*}[t]
\centering
\includegraphics[width=\linewidth]{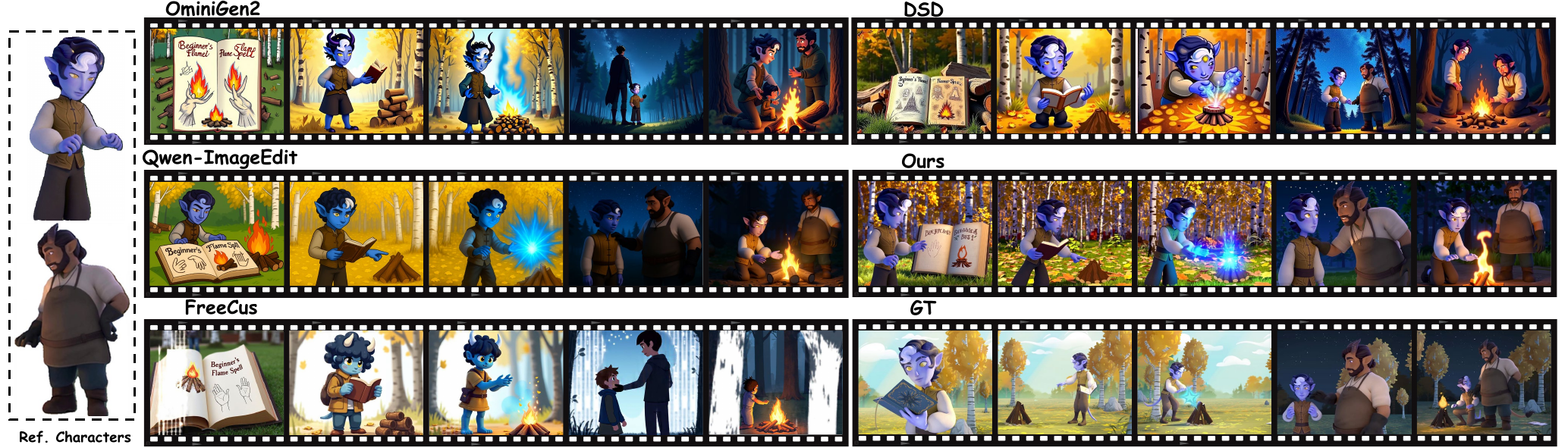}
\caption{\label{fig9}Qualitative Comparison Across Methods on Magic Instruction.}
\end{figure*}

\textbf{Qualitative results.} We further present qualitative comparisons in Figs.~\ref{fig7}-\ref{fig9}. Overall, AnimeAgent follows the script more faithfully and produces more coherent and expressive storytelling. Specifically,  Fig.~\ref{fig7} shows results on the Alice story. Anim-Director and StoryGen exhibit character drift (Alice’s outfit), Qwen-ImageEdit produces visually appealing yet logically inconsistent interactions, and FreeCus/DSD show repetitive “copy-paste” backgrounds and rigid poses. In contrast, AnimeAgent preserves character consistency and matches the narrative arc with correct key poses (bending into the hole, reaching for the door) and deliberate shot progression (close-up $\rightarrow$ wide $\rightarrow$ detail). Fig.~\ref{fig8} shows results on the Aladdin story. Morphic Studio and Aibrm suffer from identity drift (\eg inconsistent attire), GPT-4o and Nano Banana generate incorrect costume colors, and Dou bao confuses character identities. AnimeAgent maintains consistent appearances (\eg the Sorcerer’s red robe and the princess’s gray dress) and conveys the shift from trust to solemnity through coherent staging and composition. Fig.~\ref{fig9} shows results on the “magic instruction” scene. DSD and OmniGen2 miss critical visual cues, while Qwen-ImageEdit and FreeCus are inconsistent in expressions and atmosphere. Compared with GT, AnimeAgent is the closest match, preserving key



\textbf{User study.} To evaluate our results, we conducted a user study with 20 annotators (10 female, 10 male), who independently rated the generated outputs on three dimensions (\ie Character Consistency, Prompt Alignment, and Aesthetic Quality). using a 5-point Likert scale (0–4) \cite{zhuang2025vistorybench}. The stimuli comprised outputs from 20 stories randomly sampled from the VistoryBench. The average ratings are as follows: \ding{182} \textbf{Consistency}: AnimeAgent 3.95, Doubao 3.73, GPT-4o 3.42; \ding{183} \textbf{Prompt Alignment}: AnimeAgent 3.70, GPT-4o 3.54, Doubao 3.36; \ding{184} \textbf{Aesthetic Quality}: AnimeAgent 3.42, Doubao 3.34, GPT-4o 3.21.  Overall, AnimeAgent ranks best across all three dimensions, reflecting stronger character consistency, prompt alignment, and visual quality.


\subsection{Ablation Studies}

\begin{figure}[t]
\centering
\includegraphics[width=\linewidth]{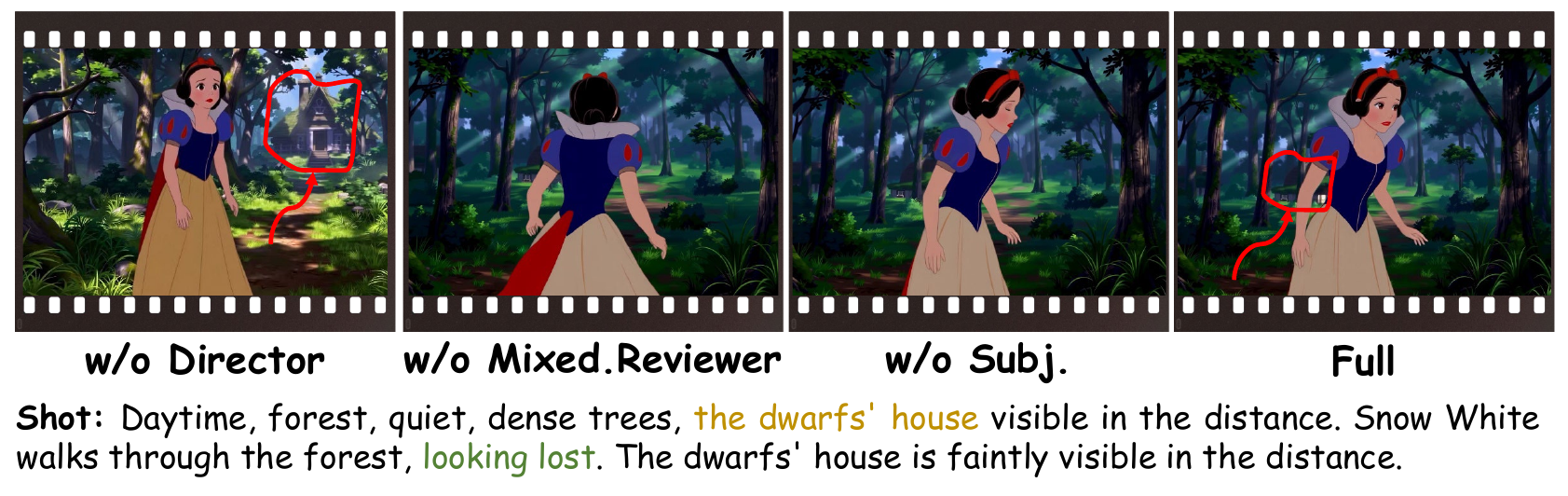}
\caption{\label{fig10}Visual Ablation Study Results.}
\end{figure}
To evaluate AnimeAgent’s components, we conduct ablation studies on ViStoryBench (Table \ref{tab4}). 

Removing the Director degrades all metrics, confirming that parsing user prompts and reference images into a structured textual dope sheet is essential for coherent motion generation. This structure provides strong semantic priors. As shown in Fig.~\ref{fig10}, w/o Director, character attire colors become inconsistent, and the house is rendered large, violating narrative logic, revealing failure to maintain visual and contextual fidelity without structured guidance.

Using Qwen3VL-30B as the Director outperforms Qwen3VL-2B, suggesting larger VLMs better capture character--scene structure. Replacing the Artist with the static T2I baseline DSD lowers performance, supporting the benefit of animator-inspired temporal generation for story clarity. Adding the Consistency Reviewer enables iterative correction of identity drift and layout errors. Without the mixed Reviewer, the final I2V frames fail to convey the narrative (Fig.~\ref{fig10}); ablating its components shows both are necessary: removing subjective evaluation reduces PA by 0.13 and weakens facial expressiveness, while removing objective metrics decreases CIDS by 0.33 and harms consistency, reflected in flatter and inaccurate expressions (Fig.~\ref{fig10}).


\begin{table}[t]
\centering
\small
\setlength{\tabcolsep}{1.0pt}  %
\renewcommand{\arraystretch}{1.1}
\begin{tabular}{@{}c|c|ccc@{}}
\toprule
\textbf{Role} & \textbf{Variant} & \textbf{CSD}$\uparrow$ & \textbf{CIDS}$\uparrow$ & \textbf{PA}$\uparrow$  \\
\midrule
\multirow{2}{*}{Module} 
  & w/o Director     & 0.581 & 0.614 & 2.75  \\
  & w/o Consi.Reviewer & 0.657 & 0.673 & 3.22  \\
\midrule
\multirow{2}{*}{Director} 
  & Qwen3VL-30B      & 0.713 & 0.766 & 3.65  \\
  & Qwen3VL-2B       & 0.689 & 0.716 & 3.43  \\
\midrule
Artist & DSD \cite{cai2025diffusion} & 0.523 & 0.664 & 3.09  \\
\midrule
\multirow{2}{*}{Mixed Reviewer} 
  & w/o Subj.        & 0.674 & 0.725 & 3.48 \\
  & w/o Obj.        & 0.661 & 0.714 & 3.41 \\
\midrule
\multicolumn{2}{c|}{\textbf{Full}} 
                     & \textbf{0.701} & \textbf{0.747} & \textbf{3.61} \\
\bottomrule
\end{tabular}
\caption{Ablation study on ViStoryBench.}
\vspace{-12pt}
\label{tab4}
\end{table}

\section{Conclusion}


We present AnimeAgent, the first I2V-based multi-agent framework for Custom Storyboard Generation (CSG). By generating trajectory-aware \textit{EXTREMES}, using a dual dope-sheet for structured guidance, and a mixed Reviewer for iterative refinement, AnimeAgent addresses the key limitations of static T2I pipelines. We also introduce a human-annotated CSG benchmark with ground-truth storyboards. Experiments show state-of-the-art performance in consistency, prompt fidelity, and animation expressiveness.

\section*{Limitations}



While AnimeAgent achieves strong performance, it depends on large MLLM and I2V backbones, leading to notable compute and latency overhead in multi-round interactions. In future work, we will investigate more efficient MLLM/I2V variants and lightweight designs to reduce inference cost and support real-time deployment.


\bibliography{custom}

\end{document}